\RequirePackage{fix-cm}

\documentclass[onecolumn,a4paper]{IEEEtran}

\usepackage{amssymb}
\usepackage{subfigure}
\usepackage{graphicx}
\usepackage{grffile}
\usepackage{amssymb}
\usepackage{color}
\usepackage{stackengine}
\usepackage{diagbox}
\usepackage{amsmath}
\usepackage{mathtools}
\usepackage{multirow}
\usepackage{pdflscape}
\usepackage{rotating}
\usepackage{float}
\usepackage{array}
\newcolumntype{P}[1]{>{\centering\arraybackslash}p{#1}}

\usepackage{adjustbox}

\usepackage[table]{xcolor}
\usepackage{collcell}
\usepackage{hhline}
\usepackage{pgf}
\usepackage{multirow}

\def\colorModel{hsb} 

\newcommand\ColCell[1]{
  \pgfmathparse{#1<50?1:0}  
    \ifnum\pgfmathresult=0\relax\color{white}\fi
  \pgfmathsetmacro\compA{1}      
\pgfmathsetmacro\compB{#1/60} 
\pgfmathsetmacro\compC{1}      
  \edef\x{\noexpand\centering\noexpand\cellcolor[\colorModel]{\compA,\compB,\compC}}\x #1
  } 
\newcolumntype{E}{>{\collectcell\ColCell}m{0.8cm}<{\endcollectcell}}  
\newcommand*\rot{\rotatebox{90}}

\usepackage{ragged2e}

\begin{document}

\title{Multi-View Region Adaptive Multi-temporal DMM and RGB Action Recognition}

\author{Mahmoud Al-Faris, John P. Chiverton, Yanyan Yang and David Ndzi}

\date{Received: data/ Accepted: date}
\maketitle

\begin{abstract}
Human action recognition remains an important yet challenging task. This work proposes a novel action recognition system. It uses a novel Multiple View Region Adaptive Multi-resolution in time Depth Motion Map  (MV-RAMDMM) formulation combined with appearance information. Multiple stream 3D Convolutional Neural Networks (CNNs) are trained on the different views and time resolutions of the region adaptive Depth Motion Maps. Multiple views are synthesised to enhance the view invariance. The region adaptive weights, based on localised motion, accentuate and differentiate parts of actions possessing faster motion. Dedicated 3D CNN streams for multi-time resolution appearance information (RGB) are also included. These help to identify and differentiate between small object interactions. A pre-trained 3D-CNN is used here with fine-tuning for each stream along with multiple class Support Vector Machines (SVM)s. Average score fusion is used on the output.
The developed approach is capable of recognising both human action and human-object interaction. Three public domain datasets including: MSR 3D Action, Northwestern UCLA multi-view actions and MSR 3D daily activity are used to evaluate the proposed solution. The experimental results demonstrate the robustness of this approach compared with state-of-the-art algorithms.\\
\textbf{keywords}:Action Recognition, DMM, 3D CNN, Region Adaptive
\end{abstract}

\section{Introduction}
\label{introduction}

Action recognition methods can be based on a number of different sources of features such as space time interest points \cite{schuldt2004recognizing}, improved trajectories of features and fisher vectors \cite{wang2013action}, \cite{perronnin2010improving}. These techniques model motion in video data which is obviously an important source of information that can be used to help recognise actions. Instead of points of motion, less localised sources of motion can also be considered to model the motion of the body as a whole such as Motion History Images (MHI)s \cite{al2017appearance} and for the boundary as with Motion Boundary Histograms (MBH)s \cite{wang2013action}. Depth can also be incorporated with techniques such as Depth Motion Maps (DMM)s \cite{yang2012recognizing}.
These sources of, what might be considered \textit{hand crafted features} are rich in information but not necessarily always able to capture all the relevant aspects of motion that might be needed to help a classifier to distinguish between different actions. 

The introduction of deep learning techniques such as Convolution Neural Networks (CNN)s \cite{krizhevsky2012imagenet} presented significant advantages for many machine learning applications, not least computer vision including action recognition, see e.g. \cite{ji20133d}. 
Deep learning based features extracted using e.g. CNNs have shown great performance over many traditional hand-crafted features due to, \textit{in simple terms} their capability to learn the important aspects of actions from the huge amount of variation that can potentially occur in images and video sequences. This property has also enabled deep learning based techniques to have improved invariance to e.g. pose, lighting, and surrounding clutter \cite{jing20173d}. It can also be seen that the inherent structure of CNN based techniques enable the preservation of the important relations in both the spatial and temporal dimensions \cite{varol2018long}.

As a part of this success, many variations in the architecture and approaches have been proposed.
A number of techniques process single video frames as static CNN features \cite{simonyan2014two,karpathy2014large}. Others \cite{simonyan2014two,wang2015towards,feichtenhofer2016convolutional} have processed short video clips where video frames were employed as multi-channel inputs to 2D CNNs. A further development is the use of 3D CNNs where Ji et al. in 2013 \cite{ji20133d} used 3D convolutions to incorporate both the spatial and temporal information of actions in video.

An extension to the conventional single stream CNN model was proposed for the first time by Simonyan and Zisserman in 2014 \cite{simonyan2014two} for action recognition. It used a two stream approach to learn single frame appearance information in combination with stacked optical flow of multiple frames which yielded improved performance. 

More recently, deep learning techniques have increasingly been used to utilise temporal information for action recognition tasks. A unique architecture was proposed in \cite{donahue2015long} using a long-term recurrent CNN with both RGB and optical flow inputs.

Temporal periods over which temporal information is learned and recognised can be very short e.g. 2 frames as in \cite{taylor2010convolutional}. Incorporating more temporal information can help improve action recognition performance, as shown by e.g. \cite{ji20133d,karpathy2014large,tran2015learning}; and multi-temporal resolution, as used by \cite{varol2018long}. These methods utilised a range of different features but the advantage of the multi-temporal resolution approach is the ability to adapt to different actions carried out at different speeds.

A deeper 3D CNN network called C3D was built in \cite{tran2015learning} and the learned motion features used different massive public video datasets. The features were shown to be compact and efficient as well as providing superior performance. The C3D model included eight convolution layers, five pooling layers, two fully connected layers.  

In \cite{yang2014dmm}, a DMM-pyramid architecture was used to train both a traditional 2D CNN and 3D CNN to keep the partial temporal information of depth sequences for action recognition. The experiments achieved comparable results with state-of-the-art methods in terms of a number of different datasets.

A CNN model obtained from ImageNet was used in \cite{wang2016action}. It was used to learn from multi-view DMM features for action recognition where a video was projected onto different view-points within the 3D space. Different temporal scales were then used from the synthesised data to constitute a spatio-temporal pattern of an action. Finally, three fine-tuned models were employed independently on the resulting DMMs. However, a fixed number of temporal scales of DMM still made the spatio-temporal information limited to action sequences carried out over a limited range of time. This would also equally need more spatio-temporal information in order for it to be recognised. In addition, some actions included object interactions which might be very difficult to discern purely from raw depth data. 

In \cite{latah2017human} a 3D CNN structure was designed to capture spatio-temporal features for action recognition. A Support Vector Machine (SVM) classifier was then used to classify actions based on the captured features. 
Experimental results showed some competitive results on the KTH action recognition data.
Similarly, a 3D CNN was proposed in \cite{baccouche2011sequential} to automatically extract spatio-temporal features. Then however, a Recurrent Neural Network (RNN) was used to classify each sequence considering the learned features for each timestep. The experiments on the KTH dataset demonstrated impressive performance in comparison to state-of-the-art approaches. Another use of a 3D CNN was by Taylor et al. in 2010 \cite{taylor2010convolutional} with a Restricted Boltzmann Machine to learn spatio-temporal features.

An efficient approach was proposed by Liu et al. in 2017 \cite{liu2017two} which used a joint-pooled 3D deep convolutional descriptor applied to skeletal feature data on action recognition data. The experimental results demonstrated promising performance.
Temporal information was exploited in \cite{Srijan2018} which used a deep Long Short Term Memory (LSTM) method on skeleton based data sequences, which was then combined using a fusion based approach with appearance information and employed for action recognition.

Deep learning based action recognition was also presented in \cite{liu20163d} using depth sequences and skeleton joint information combined. A 3D CNN structure was used to learn the spatio-temporal features from depth sequences, then Joint-Vector features were computed for each sequence. Finally, the SVM classification results of the two types of features were fused for action recognition.

The 3D positional information in depth data can be further emphasised, as was done by \cite{xiao2019action} where multiple views were derived of the depth data. The authors applied it to dynamic depth images rather than incorporating it into a DMM formulation. 

All these different sources of features are useful but most of them do not consider the way the motion might be carried out over different ranges of time. This can be considered important in cross-actor and even for the same actor at different time points or similar. Appearance information is also not commonly used. Also, little attention is given to how different image regions that might be considered of higher relevance for different actions. Furthermore, they do not consider the affect of higher level information (e.g.\ \textit{pose}) on the underlying learnt feature space. 

\begin{figure*}[ht!]
\centering
\includegraphics[width=0.9\textwidth]{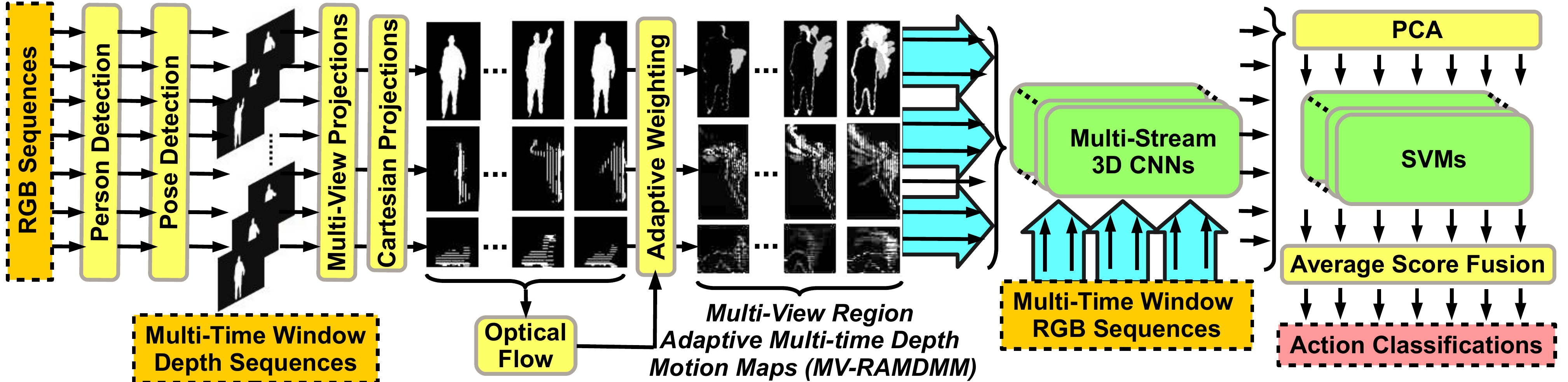}
\caption{Framework of our hierarchical Region Adaptive Multi-time resolution Depth Motion Map (RAMDMM) and Multi-time resolution RGB action recognition system. Each pose, Cartesian projection, view and time window has a separate 3D CNN and SVM. The system is configured here to detect 2 poses, across 7 views and 3 time resolutions in the 3 Cartesian planes. The RGB information is also detected across 3 time resolutions. This results in $2(\times 3\times 7 \times 3+3)=132$ separate 3D CNNs and SVM classifiers. }
\label{recognitionframework}
\end{figure*}

At a lower level, it can also be considered preferable to obtain motion information from multiple contiguous frames in addition to the spatial information. Therefore, more suitable approaches are needed to capture extra temporal information as well as to keep the complexity of the model as low as possible. To this end, we propose a new hierarchical pose detection and action recognition system. 
The pre-trained C3D model is adapted here to learn multi-resolution features from both the spatial and temporal dimensions using different contiguous frames of RGB data. Furthermore, we propose an adaptive Multi-resolution Depth Motion Map calculated across multiple views with important action information learned through the 3D CNN model to provide extra motion based features that emphasise the significance of moved parts of an action.
In addition, multi-resolution raw appearance information (i.e.\ RGB) is used to exploit various spatio-temporal features of the RGB scene which help to capture more specific information that might otherwise be difficult to obtain from depth sequence information alone such as object-interactions and finer image details. 
Our adaptive action recognition system is illustrated in Fig. \ref{recognitionframework}.

Our automated system is developed and evaluated based on three well-known publicly available datasets including the Microsoft Research (MSR) Action 3D dataset \cite{li2010action}, the North Western UCLA Multiview Action 3D dataset \cite{wang2014cross} and the MSR daily activity 3D dataset \cite{wang2012mining}. 
The experimental results demonstrate the robustness of out approach compared with state-of-the-art algorithms. 

\section{Methodology}

Traditional Depth Motion Maps (DMM)s are formulated on 2D planes by combining projected motion maps of an entire depth sequence. This does not consider the higher order temporal links between frames of depth sequences. 
A DMM can encapsulate a certain amount of the variation of a subject's motions during the performance of an activity. Unfortunately difficulties can arise for activities that have the same type of movements but performed over different temporal periods. Our formulation therefore includes multiple time resolutions, referred to as Multi-resolution DMM (MDMM).
Moreover, some actions or parts of actions are performed with different intensities. The differences in depth information captured at points of fast motion is accentuated using a region and motion adaptive formulation producing a Region Adaptive MDMM (RAMDMM). This adaptivity helps to further differentiate between actions, particularly with differences in depth due to positioning compared with actions with fast motion.

\subsection{Depth Motion Maps}

The basic DMM, (as used in e.g. \cite{yang2012recognizing,chen2016realsensor,chen2016real}), includes projecting each depth frame onto three orthogonal Cartesian planes.
The motion energy from each projected view is then stacked. This can be through a specific interval or through the entire sequence to generate a Depth Motion Map (DMM), $\Gamma_v$ for each projection view,
\begin{equation}
    \Gamma_v(t) = \sum_{t'=t}^{t+N-1} | m_v^{{t'}+1} - m_v^{t'} |
    \label{DMM}
\end{equation}
where $v \in$ \big\{ xy, yz, xz \big\} indicates the Cartesian projection; $m_v^t$ is the projected map of the depth information at time frame $t$ under projection view $v$; $N$ is the number of frames that indicates the length of the interval.
DMMs can be represented by combining the three generated DMMs $\Gamma_v$ together where important information of body shape and motion are emphasised. Average score fusion is used here, to be discussed shortly in section \ref{sec:classification}.

\subsubsection{Multi-resolution-in-time Depth Motion Maps}

Mostly, a fixed number of frames have been used by other researchers or even the entire number of frames of an action sequence video to generate DMMs. But a length of an action is not known in advance. Hence, multi-resolution-in-time depth motion maps are needed to cover different temporal intervals and rates of an action. 

To produce a Multi-resolution DMM (MDMM), the depth frames from a depth sequence are combined across three different ranges where each has a different time interval. This means that various values of temporal length, $\tau$ are set to generate the MDMMs for the same action (depth sequence). As $\tau$ $\in$ ${N^+}$ in traditional DMMs, this can be improved by $\tau$ $\in$ $\big\{\lambda_{1}, \lambda_{2}, \lambda_{3} \big\}$ where $\lambda_{i}$ $\in$ $N^+$ are different temporal windows used to properly cover an action's motion regardless of whether it carries important information over a short or long duration. Each of these three duration's produce a different DMM. The values of $\tau$ are selected to cover short, intermediate and long duration's, where $long$ would typically correspond to an entire depth sequence for the various video sequences considered here.

These MDMMs for each depth sequence can be calculated with:
\begin{equation}
    \Gamma_{v,\tau}(t) = \sum_{t'=t}^{t+\tau-1} |m^{{t'}+1}_v - m^{t'}_v| 
\end{equation}
where $v \in \big\{xy, yz, xz\big\}$, $\tau$ $=$ $\lambda_i$ and, e.g. $\tau \in \big\{5, 10, All\big\}$ (as used here) are the various lengths of depth sequence used to obtain a MDMM for each single frame.

\subsubsection{Adaptive Motion Mapping}

As already considered, different actions can be performed over different time periods. The MDMM is able to include motion information across a range of temporal windows. However each action can also be performed at different speeds by different people and with movement in different locations in an image.
Hence, an adaptive weighting approach based on the movement is applied to continuously weight the interest regions to adapt to any sudden change in an action.

To adapt various changes of an action, an adaptive weighting approach based on the magnitude of the optical flow motion vectors is employed to build a regional adaptive MDMM. Firstly, motion flow vectors are extracted using optical flow as explained in \cite{liu2008human} on consecutive frames. Then the motion magnitude for each single pixel is computed and normalised between two consecutive frames. 

\begin{figure}[th]
\centering
\includegraphics[width=0.49\textwidth]{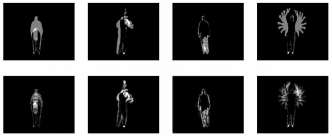}
\caption{Samples of traditional (top row) and adaptive weighted (bottom row) DMM templates (left to right): Bend, Tennis serve, Forward kick and Two hands wave.}
\label{DMMtypes}
\end{figure}

Optical flow is computed between two consecutive frames, i.e. $I_t$, and $I_{t+1}$. The result of the optical flow function is the motion flow vector $\boldsymbol{o}$ with vector elements $\boldsymbol{o}_x$ and $\boldsymbol{o}_y$ in the vertical and horizontal directions respectively. The motion magnitude of the flow vectors of each pixel can be calculated using:
$g = \boldsymbol{o}_x^2+\boldsymbol{o}_y^2$.
As the motion magnitude changes based on the type, speed and shape of an action movement, this can be utilised to improve the DMM calculation formula by including the motion magnitude in the DMM equation as a weighting function. This helps to add increased consideration for higher interest regions of a DMM template as well as providing low consideration for other regions. In addition, it can make the DMM template adapt to different movements in an action movement. The new RAMDMM can be formulated as follows:
\begin{equation}
    \Gamma^{\rm of}_{v,\tau}(t) = \sum_{t^{'}=t}^{t+\tau-1} \left(|m^{t^{'}+1}_v - m^{t'}_v| \times g^{t'+1}_v\right) 
\end{equation}
where $g_{v}^{t'+1}$ is the motion magnitude for view $v$ at time point $t'+1$.
Fig. \ref{DMMtypes} shows samples of DMM templates illustrating some differences between traditional DMMs and the region adaptive DMM method.

\subsection{Multiple Views}
The 3D characteristic of the depth sequences mean that it is possible to calculate different view points of the same data. This can help to improve the model by making it view invariant.
A virtual camera can be rotated with a specific value in 3D space, which can be seen to be equivalent to rotating the 3D points of the depth frames.

The virtual camera can be moved within the depicted space, for instance, from point $\boldsymbol{p}$ to $\boldsymbol{p}'$. The first step is to move from $\boldsymbol{p}$ to $\boldsymbol{p}_b$ with rotation angle $\alpha$ around Y-axis, then from $\boldsymbol{p}_b$ to $\boldsymbol{p}'$ with rotation angle $\beta$ around X-axis. This is performed by the rotation matrices:
\begin{equation}
    \mathbf{R}_x = 
    \begin{bmatrix}
    \cos(\beta) & 0 & \sin(\beta)\\
    0 & 1 & 0\\
    -\sin(\beta) & 0 & \cos(\beta)
    \end{bmatrix},
\end{equation}
and
\begin{equation}
    \mathbf{R}_y = 
    \begin{bmatrix}
    1 & 0 & 0\\
    0 & \cos(\alpha) & -\sin(\alpha)\\
    0 & \sin(\alpha) & \cos(\alpha)
    \end{bmatrix},
\end{equation}
The right-handed coordinate system is used for the rotation where the original camera view-point is $\boldsymbol{p}$. Hence, the new coordinate of 3D point after rotation can be considered as follows:
\begin{equation}
    \boldsymbol{p}' = \mathbf{R}_x \mathbf{R}_y \boldsymbol{p}
\end{equation}
where $\boldsymbol{p}'$ is the new coordinate, and the corresponding depth value for the synthesised depth frames. 

\begin{figure}[th]
\centering
\includegraphics[width=0.49\textwidth]{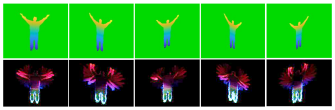}
\caption{Samples of original and synthesised depth frames ($1^{st}$ row) with RA-DMM ($2^{nd}$ row) after multiple view-points rotation when (left to right): $\alpha$ $\in$ (0, 30, -30, 45, -45) respectively.}
\label{viewsamples}
\end{figure}

Our view projection method on depth sequences is similar to \cite{xiao2019action} except applied here to enable extraction of DMMs. Some results of multi-view projection are presented in Fig. \ref{viewsamples} with different values of $\alpha$ rotation angles. It can be noticed that more discriminative information can be obtained by computing RA-DMM based on the synthesised depth frames. 

Sequences of synthesised depth frames with different view-points can be synthesised from a series of these multi-view projections. This can contribute to better data augmentation for training processes in addition to better overall feature extraction.

In terms of the DMM formulation, multiple views extends the formulation with an additional dependency term, i.e.
\begin{equation}
    \Gamma^{\rm of}_{v,\tau,\alpha}(t) = \sum_{t^{'}=t}^{t+\tau-1} \left(|m^{t^{'}+1}_{v,\alpha} - m^{t'}_{v,\alpha}| \times g^{t'+1}_{v,\alpha}\right) 
\end{equation}
where $\alpha\in\mathcal{A}$ is a sequence of angular values where $\mathcal{A}=(-45,...,45)$. Here $\mathcal{A}=(-45,-30,-15,0,15,30,45)$ so that $|\mathcal{A}|=7$.

\subsection{Feature Extraction, Classification and Fusion}
\label{sec:classification}

An effective approach was presented for action recognition in \cite{tran2015learning} to learn spatio-temporal features using a 3D convolutional neural network which was also trained on a number of different large video datasets. The training settings were kept the same as the original C3D model. 

A 3D CNN is able to capture temporal information based on 3D convolution and pooling operations which are performed in the spatial and temporal dimensions.

The C3D network has eight convolution layers and five pooling layers that followed on from each other. Two fully-connected layers and a softmax loss layer are used to recognise at the individual action label level. The number of kernels are 64, 128, 256, 256, 512, 512, 512, 512 for the convolution layers. The size of all kernels in the 3D CNN was set to 3 $\times$ 3 $\times$ 3 with stride 1 $\times$ 1 $\times$ 1. For the 3D pooling layers, the kernel sizes were set to 2 $\times$ 2 $\times$ 2 with stride 2 $\times$2 $\times$ 2 except for the first pooling layer which had a kernel size of 1 $\times$ 2 $\times$ 2 and a stride of 1 $\times$ 2 $\times$ 2 in order to preserve the temporal information at the early stages. The fully connected layers have 4096 output units each.

Conventionally, the value at position $(x, y, z)$ on the $j$th feature map in the $i$th layer can be formulated as follows:
\begin{align}
    \zeta_{ij}&^{xyz}=\\
&\tanh\Big(b_{ij}\hspace{-1mm}+\hspace{-1mm}\sum_m \sum_{p=0}^{P_{i}-1} \sum_{q=0}^{Q_{i}-1} \sum_{r=0}^{R_{i}-1} w_{ijm}^{pqr} \zeta_{(i-1)m}^{(x+p)(y+q)(z+r)}\Big),\nonumber
\end{align}
where $\tanh(.)$ is the hyperbolic tangent function, $b_{ij}$ is the bias for this feature map, $m$ indexes over the set of feature maps in the $(i-1)$th layer connected to the current feature map, $w_{ijm}^{pqr}$ is the value at the position $(q, p, r)$ of the kernel connected to the $m$ feature map in the previous layer. The kernel sizes, $R_i$, $P_i$ and $Q_i$ are the temporal and spatial (height and width) dimensions, respectively. 

Each value of the Cartesian projections $v$, time resolution $\tau$ and view $\alpha$ has a separate 3D CNN model that is trained based on a set of actions. The 2D output of each 3D CNN is then split into temporal feature vectors and concatenation of the three orthogonal views is used to form a single feature vector. The dimensionality of each resulting feature vector is then reduced using Principal Component Analysis (PCA) determined from a covariance matrix of all the feature vectors. The projected feature vectors are then fed into different multiple class Support Vector Machines (SVM)s \cite{FAN_ET_AL_2008} that are trained to recognise actions.
The 3D CNN is trained to use a fixed number of input frames ($\lambda=16$) for the depth information, i.e.
\begin{align}
  &\mbox{CNN}^{\lambda}_{v,\tau,\alpha}(t)\\
&=\mbox{CNN}_{v,\tau,\alpha}
\left(\Gamma'_{v,\tau,\alpha}(t),\Gamma'_{v,\tau,\alpha}(t-1),...,\Gamma'_{v,\tau,\alpha}(t-\lambda)\right)\nonumber
\end{align}
where $\Gamma'_{v,\tau,\alpha}(t)$ is a scaled and colour mapped (jet) version of the multi-view region adaptive multi-temporal resolution feature data $\Gamma^{\rm of}_{v,\tau,\alpha}(t)$ for time $t$.
The input frame size of the pre-trained C3D network is also fixed. A padding technique and interpolation are used here to resize frames to the required dimensions. Following the 3D CNN feature extraction process, feature concatenation and dimensionality reduction, the SVM classification is performed:
\begin{equation}
\boldsymbol{c}_{v,\tau,\alpha}(t)=\mbox{SVM}_{v,\tau,\alpha}
\left(\mbox{CNN}^{\lambda}_{v,\tau,\alpha}(t)
\right).
\end{equation}
Classification vectors are then combined across all Cartesian planes, resolutions and views using average score fusion of the form:
\begin{equation}
\boldsymbol{c}^{\rm dmm}(t)=\frac{1}{|V\times\Lambda\times\mathcal{A}|}\sum\limits_{v}\sum\limits_\tau\sum\limits_\alpha\boldsymbol{c}_{v,\tau,\alpha}(t).
\end{equation} 

\subsection{Multi-Resolution Spatio-Temporal RGB Information}

Some types of actions and motions especially those that interact with objects can be perceived better with appearance information rather than e.g. depth due to the differences in the characteristics of the object in terms of appearances. In addition, it is somehow difficult to capture the DMM information of these objects especially when the object's state is fixed or the size is relatively small.

Therefore, RGB data is utilised in this work as a source of the appearance information within our 3D CNN network model to capture discriminative spatio-temporal information of both subjects and interacting-objects. Moreover, different temporal scales are used to cover different temporal ranges in the RGB scene, the same as for RAMDMM. This can help to mitigate against problems that might arise due to variations in the speed at which actions are performed that could result with different action performers. Three temporal scales 
are employed across three independently fine-tuned C3D models (in fixed mode for $\lambda=16$ but then updated to use a variable number of inputs with $\lambda\in\{10,25\}$); the outputs of which are fed into three independently trained multi-class Support Vector Machines (SVM)s. The outputs of the SVM classifiers are then combined together via average score fusion to form the multi-resolution RGB information:
\begin{equation}
    \boldsymbol{c}^{{\rm rgb}}(t) = \frac{1}{|R|}\sum\limits_{r\in R}\boldsymbol{c}^{{\rm rgb}}_{r}(t)
\label{mrrgb}
\end{equation}
where $\boldsymbol{c}^{{\rm rgb}}_{r}$ is the action classification vector for the RGB image frames taken across a time window of $r$ frames.

An overall average score fusion is then used to derive the final classification vector, given by
\begin{equation}
\boldsymbol{c}(t)=\tfrac{1}{2}\left(\boldsymbol{c}^{\rm rgb}(t)+\boldsymbol{c}^{\rm dmm}(t)\right).
\end{equation}

\subsection{People Detection and Pose Classification}

The action recognition system can be made to perform well across a wide range of actions however this task can be further enhanced if the person performing an action can be localised in the image space. This helps remove extraneous background clutter and distractors. The performance of the system can also be further enhanced if the pose of the person can be detected prior to action recognition. It can be considered that this would help to provide the classification system with a better defined delineation between different actions performed in different poses. For instance, using a telephone whilst standing or sitting could produce a range of features that may not be that well connected in feature space or separated from other features from other actions.

Person detection is performed here using the Faster R-CNN \cite{ren2017faster} person detector, based on the AlexNet \cite{NIPS2012_4824} model as a network structure but transformed into a Region Proposal Network (RPN), with the use of a ROI max pooling layer and classification layers. 

A few samples from the RGB data of the utilised datasets are used to create the ground-truth training data. After training, the created Faster R-CNN network is then used for person detection on the RGB data. This can help to eliminate the noise of the background environment in the action recognition process as can be seen in section \ref{results}.

Pose detection is performed here using a specially adapted AlexNet pre-trained model \cite{krizhevsky2012imagenet} using transfer learning to classify the pose of an occupant out of three specified poses ($\bf{sitting}, \bf{standing}, \bf{laying}$).

\section{Experiments \& Results} 
\label{results}

Three public datasets are used to evaluate the proposed method for action recognition: North Western UCLA Multiview Action 3D dataset \cite{wang2014cross}; Microsoft Research (MSR) Action 3D dataset \cite{li2010action}; and the MSR daily activity 3D dataset \cite{wang2012mining}. 

The overall steps and parameter values that are employed on the datasets for feature extraction and action recognition are summarised as follows:
\begin{itemize}
    \item Project the original depth sequence into different views with $\alpha$ $\in$ $(45,30,15,$ $-15,-30,-45)$ which results in 6 synthesised views of the data and the original at $\alpha=0$;
    \item Compute Cartesian projections of the 7 views;
    \item Compute motion vectors' magnitude using optical flow algorithm over the original and synthesised sequences;
    \item Compute RA-DMMs for each sequence of original and synthesised sequences;
    \item Each action sequence is split into 16 frame sub-sequences in terms of RA-DMMs and (10,16,25) frames sub-sequences for the RGB information to train the 3D CNNs;
    \item Compute RA-DMM of each sequence of original and synthesised sequences using sub-sequence concatenation, dimensionality reduction then average score fusion of three maps templates;
    \item Multi-resolution RA-DMM computed using average score fusion for RA-DMM windows $(5,10,\mbox{all})$;
    \item Multi-view RAMDMM computed using average score fusion of SVM classifiers for all RAMDMM across different views;
    \item Multi-resolution RGB (depth in MSR 3D action dataset) information computed using average score fusion of SVM classifiers for $\{10, 16, 25\}$ frames;
    \item Overall proposed system achieved with average score fusion between MV-RAMDMM and MR-RGB.
\end{itemize}

\subsection{North Western UCLA Dataset}

North Western UCLA Multiview action 3D dataset \cite{wang2014cross}
has three Kinect cameras used to capture RGB, depth and human skeleton data simultaneously. This dataset includes 10 different action categories including: pick up with one hand, pick up with two hands, drop trash, walk around, sit down, stand up, donning, doffing, throw, carry. Each action is performed by 10 actors. In addition, this dataset consists of a variety of viewpoints.

We evaluate our proposed method with two different training and testing protocols for this dataset:
\begin{itemize}
    \item Cross-subject training scenario: In this setting we use the data of 9 subjects as training data, and leave the data of the remaining subject as test data. This is useful to show the performance of the recognition system across subjects. Furthermore, this is a standard criteria for comparison with the state-of-the-art.
    \item Cross-view training scenario: As this dataset contains three view cameras, we use the data of 2 cameras as training data, and leave the remaining camera as test data. This kind of setting is used to demonstrate the ability of the recognition system to perform with different views and to get another standard criteria to compare with the state-of-the-art.
\end{itemize}

These settings give the opportunity to evaluate the proposed system with variations for different subjects and different views. The proposed method achieves an interesting set of results for the complete system demonstrating state-of-the-art performance as can be seen shortly. But first, let us examine the performance of the individual streams with individual inputs.

\subsubsection{Multi-Resolution in Time Appearance Information}
To start, the classification performance using multi-temporal resolution RGB data as an input to the 3D CNN model (C3D) is investigated together with the multi-class SVM classifier based on the aforementioned evaluation scenarios. Three temporal resolutions are used in terms of the RGB model including ${10, 16, 25}$ windows. The trainable layers are adapted in the 3D CNN model when a non-conformant input is used, i.e. $\lambda\in{10, 25}$. An average score fusion is employed between the three SVMs to produce the multi-temporal resolution of the RGB data for action recognition. Table \ref{C3Dtable} includes the results for the different temporal resolutions for the RGB data in addition to the average score fusion result.

\begin{table} [th]
\centering
\caption{Results of the fine tuned C3D model with a multi-class SVM classifier for different time resolutions of the RGB data for the North Western UCLA dataset. \label{C3Dtable}}
\begin{tabular}{|c|c|c|c|c|}
 \hline
 {Settings}&{RGB$_{10}$} & {RGB$_{16}$} & {RGB$_{25}$} & {RGB$_{\rm fusion}$}\\
 \hline
 Cross-subject&67.44&78.12&88.23&91.51\\
 \hline
 Cross-view&56.71&61.79&70.32&72.20\\
\hline 
\end{tabular}
\end{table}

As we can see in Table \ref{C3Dtable}, the fine-tuned C3D model achieves good performance in terms of cross-subject and cross-view classification schemes. The model already achieves relatively good recognition rates particularly as the temporal window increases. It can be seen that C3D with multi-class SVM classification on RGB data alone with 25 temporal frames achieves the highest recognition performance of 88.23\% and 70.32\% in terms of cross subject and view evaluation schemes respectively. This reduces to (78.12\%, 61.79\%) and (67.44\%, 56.71\%) when 16 and 10 temporal frames are used respectively. Finally, the highest overall recognition performance is achieved when average score fusion is employed, combining the outputs of the three temporal results, again for RGB scene information only.

\subsubsection{Multi-Resolution in Time Region Adaptive Depth Motion Maps}
The Region Adaptive DMM (RADMM) templates are calculated across the three temporal resolutions to form the multi-resolution DMM template, referred to as RAMDMM. These are used to learn discriminative features encapsulating depth, time and motion information. Results demonstrating the improvements achieved for the depth across multiple time windows are shown in Table \ref{ramdmm}.
\begin{table} [th]
\centering
\caption{Results of the proposed model used RADMM and RAMDMM templates in terms of the North Western UCLA dataset.\label{ramdmm}}
\begin{tabular}{|c|c|c|c|c|}
 \hline
 {Settings}&{5} & {10} & {All} & {RAMDMM}\\
 \hline
 Cross-subject&79.14&86.10&91.32&93.87\\
 \hline
 Cross-view&61.20&67.22&75.95&77.15\\
\hline 
\end{tabular}
\end{table}
A similar trend as was seen for the appearance information can be observed for these results, i.e. a greater time window increases recognition performance which is further improved by average score fusion for all time windows combined.

\subsubsection{Combining RAMDMM, Multiple Views and Appearance Based Multiple Sequences}
The depth, time and motion information are then further combined across multiple synthesised views to produce MV-RAMDMM based action recognition. At the end, an average score fusion is employed between the MR-RGB and MV-RAMDMM to utilise appearance, motion, shape, and historical information based action recognition. Table \ref{ramdmm2} includes the results of the proposed method at different stages in the action recognition.
\begin{table}[th]
\centering
\caption{Results of the proposed model used RAMDMM, MV-RAMDMM templates and average score fusion with MR-RGB in terms of the North Western UCLA dataset. \label{ramdmm2}}
\begin{adjustbox}{max width= 0.49 \textwidth}
\begin{tabular}{|c|c|c|c|c|}
 \hline
 {Settings}& {\tiny RAMDMM} & {\tiny MV-RAMDMM} & {\tiny MV-(RAMDMM+RGB)} \\
 \hline
 Cross-subject&93.87&96.30&97.15\\
 \hline
 Cross-view&77.15&84.52&86.20\\
\hline 
\end{tabular}
\end{adjustbox}
\end{table}
The results in Table \ref{ramdmm2} appear to show that the different views of RAMDMM encapsulated within the MV-RAMDMM streams help to significantly improve the recognition rate for both the cross-subject and cross-view settings.
In addition, an average score fusion between MV-RAMDMM and MR-RGB gives the opportunity to share a variety of important information for action recognition; improving the recognition accuracy in comparison to individual model classification reaching to 97.15 \% and 86.20 \% in terms of cross-view and cross-subject evaluation schemes respectively. This can be compared to state-of-the-art approaches as seen in Table \ref{UCLAcomparison2}.
\begin{table} [th]
\centering
\caption{A comparison between the proposed method and state-of-the-art approaches in terms of North Western UCLA dataset. \label{UCLAcomparison2}}
\begin{tabular}{|c|c|c|}
 \hline
 {Paper}&{Cross-subject} & {Cross-view} \\
 \hline
  Virtual view \cite{li2012discriminative}&50.70&47.80\\
  Hankelet \cite{li2012cross}&54.20&45.20\\
  MST-AOG \cite{wang2014cross}&81.60&73.30\\
  Action Bank \cite{sadanand2012action}&24.60&17.60\\
  Poselet \cite{maji2011action}&54.90&24.50\\
  Denoised-LSTM \cite{demisse2018pose} & - &   79.57\\
 tLDS \cite{ding2018tensor} & 92.99 & 74.6\\
MVDI \cite{xiao2019action} & - & 84.20\\
\hline
 \bf{Ours}&\bf{\textit{97.15}}&\bf{\textit{86.20}}\\
\hline 
\end{tabular}
\end{table}

It can be seen in Table \ref{UCLAcomparison2}, Virtual view \cite{li2012discriminative} and Hanklet \cite{li2012cross} methods are limited in their performance which reflects the challenges of the North Western UCLA dataset (e.g. noise, cluttered backgrounds and various view points). To mitigate these challenges, MST-AOG was proposed in \cite{wang2014cross} and achieved 81.60\%. Our method achieves a significant improvement of 18\% over MST-AOG and some comparable performance for the cross-view setting due to the big challenge in a cross-view setting. A confusion matrix of the proposed method is shown in Figure \ref{UCLAview2nd} using spatial and motion information in terms of the North Western UCLA multi-view action 3D dataset.

\begin{figure}[htb!]
\centering
\newcommand\items{10}   
\begin{adjustbox}{max width= 0.49 \textwidth}
\begin{tabular}{cc*{\items}{|E}|}
\arrayrulecolor{white}
\multicolumn{1}{c}{} &\multicolumn{1}{c}{} &\multicolumn{\items}{c}{} \\ \hhline{~*\items{|-}|}
\multicolumn{1}{c}{} & 
\multicolumn{1}{c}{} & 
\multicolumn{1}{c}{\rot{\textbf{Carry}}} & 
\multicolumn{1}{c}{\rot{\textbf{Doffing}}} & 
\multicolumn{1}{c}{\rot{\textbf{Donning}}}&
\multicolumn{1}{c}{\rot{\textbf{Drop trash}}} & 
\multicolumn{1}{c}{\rot{\textbf{Pick up-one}}} & 
\multicolumn{1}{c}{\rot{\textbf{Pick up-two}}}&
\multicolumn{1}{c}{\rot{\textbf{Sit down}}} & 
\multicolumn{1}{c}{\rot{\textbf{Stand up}}}&
\multicolumn{1}{c}{\rot{\textbf{Throw}}}&
\multicolumn{1}{c}{\rot{\textbf{Walk around}}}\\
\hhline{~*\items{|-}|}
\multirow{\items}{*}{\rotatebox{90}{}} 
&\textbf{Carry}  & 79 & 0 & 4 & 1 & 0 & 0 & 0 & 0&10 &6 \\ \hhline{~*\items{|-}|}
&\textbf{Doffing}  & 0 & 100 & 0 & 0 & 0 & 0 & 0 & 0&0&0   \\ \hhline{~*\items{|-}|}
&\textbf{Donning}  & 0 & 6 & 94 & 0 & 0 & 0 & 0 & 0&0&0  \\ \hhline{~*\items{|-}|}
&\textbf{Drop trash}  & 5 & 0  & 0 & 77 & 12 & 0 & 0 & 0 &0 &6   \\ \hhline{~*\items{|-}|}
&\textbf{Pick up-one}  & 0 & 0 & 0 & 0 & 65 & 16 & 2 & 12 &0&5  \\ \hhline{~*\items{|-}|}
&\textbf{Pick up-two}  & 1 & 0  & 0 & 3 & 11 & 83 & 0 & 0&0&2   \\ \hhline{~*\items{|-}|}
&\textbf{Sit down}  & 0 & 0  & 0 & 0 & 0 & 0 & 100 & 0&0&0   \\ \hhline{~*\items{|-}|}
&\textbf{Stand up}  &  0 & 0  & 0 & 0 & 0 & 0 & 0 & 100 &0&0  \\ \hhline{~*\items{|-}|}
&\textbf{Throw}  & 17 & 0 & 9 & 0 & 6 & 0 & 0 & 0 &68&0  \\ \hhline{~*\items{|-}|}
&\textbf{Walk around}  & 1 & 0 & 0 & 2 & 1 & 0 & 0 &0&0&96   \\ \hhline{~*\items{|-}|}
\end{tabular}
\end{adjustbox}
\caption{Confusion matrices of the proposed method, using view-setting validation scheme in terms of North Western UCLA dataset.}
\label{UCLAview2nd}
\end{figure}

\subsection{MSR 3D Action Dataset}

The Microsoft Research (MSR) Action 3D dataset \cite{li2010action} is an action dataset consisting of depth sequences with 20 actions: high arm wave, horizontal arm wave, hammer, hand catch, forward punch, high throw, draw cross, draw tick, draw circle, hand clap, two hand wave, side-boxing, bend, forward kick, side kick, jogging, tennis serve, golf swing, pickup and throw. Each action is performed three times, each by ten subjects. A single point of view is used where the subjects were facing the camera while performing the actions. 
The dataset has been split into three groups based on complexity: AS1, AS2 and AS3 as used in many studies see e.g. \cite{li2010action,yang2012recognizing,chen2016real,xia2012view}.

The action subsets are summarised in Table \ref{subsets}. All validation schemes make use of the three subsets.
\begin{table}[th]
\centering
\caption{Subsets of MSR action 3D dataset \cite{li2010action}.
\label{subsets}}
\begin{adjustbox}{max width= 0.49 \textwidth}
\begin{tabular}{|c|c|c|}
 \hline
  {AS1} & {AS2} & {AS3} \\
 \hline
 Horizontal arm wave&High arm wave&High throw\\Hammer&Hand catch&Forward kick\\Forward punch&Draw tick&Side kick\\High throw&Draw cross&Jogging\\Hand clap&Draw circle&Tennis swing\\Bend&Two-hand wave&Tennis serve\\Tennis serve&Side-boxing&Golf swing\\Pick-up and throw&Forward kick&Pick-up and throw\\
 \hline
\end{tabular}
\end{adjustbox}
\end{table}
Three evaluation schemes are considered in the literature (see e.g. \cite{padilla2014discussion}) in terms of the MSR action 3D dataset:
\begin{itemize}
    \item 1/3 evaluation scheme: 1/3 of the instances are used as training samples and the reminder as testing samples. The 1/3 scheme splits the dataset using the first repetition of each action performed by each subject as training, and the rest for testing.
    \item 2/3 evaluation scheme: 2/3 of the instances are used as training samples and use the remainder as testing samples. The 2/3 scheme splits the dataset into training samples using two repetitions of each action performed by each subject and testing uses the rest of the data.
    \item Cross-subjects evaluation scheme: half of the subjects are used as training samples, the other half are used as testing samples. Any half of the subjects can be used for testing, e.g. 2, 4, 6, 8 and 10; and the rest for training, i.e. 1, 3, 5, 7 and 9 (as used here).
\end{itemize}
Each subset has eight actions that can be used to evaluate the proposed method in terms of 1/3, 2/3, and cross-subject validation schemes. These can help to assess the performance of the proposed method against different training settings such as shortage of training samples, many training samples and variations between different subjects.

Similar to the experiments conducted above for the North Western UCLA 3D action data set, a series of progressive sets of experiments are carried out.

\subsubsection{Depth Information}
\label{msrdepthsequences}
This data set only has depth information (no appearance information). Therefore, instead of RGB based appearance information, the depth frames are used. 
 The pre-trained C3D network is individually implemented based on depth data (instead of RGB) with various temporal frames ${10, 16, 25}$ for the different MSR evaluation schemes. Then, an average score fusion is employed between the models to show the effect on the recognition rate. Table \ref{C3Dmodelmsr} includes the results of the C3D network implementation based on depth data alone.
\begin{table}[th]
\centering
\caption{Performance of the C3D model based on multi-resolutions depth information in terms of MSR 3D Action dataset. \label{C3Dmodelmsr}}
\begin{adjustbox}{max width= 0.49 \textwidth}
\begin{tabular}{|c|c|c|c|c|c|}
 \hline
  {Subsets} & {Scheme} & {Depth$_{10}$} & {Depth$_{16}$} & {Depth$_{25}$} & {Depth$_{\mbox{fusion}}$}\\
 \hline
 \multirow{3}{*}{AS1}&1/3 &64.80&65.32&72.87&74.51\\&2/3 &75.30&76.71&77.14&80.10\\&Cross &47.82&53.81&57.20&60.20\\
   \hline 
 \multirow{3}{*}{AS2}&1/3 &58.40&61.23&67.01&71.40\\&2/3 &61.72&68.18&74.89&76.72\\&Cross &50.61&51.59&55.20&56.91\\
 \hline
 \multirow{3}{*}{AS3}&1/3 &65.23&69.10&71.60&74.82\\&2/3 &69.17&78.43&80.94&81.10\\&Cross &51.21&57.51&59.88&61.13\\
\hline
\end{tabular}
\end{adjustbox}
\end{table}
Again, the recognition performance is improved with greater temporal windows and by using a combination of different temporal dimensions combined by average score fusion; making the system more robust against speed variations. This demonstrates the utilisation of shape and temporal information from the depth sequences in the recognition process.

\subsubsection{Multi-Resolution in Time Region Adaptive Depth Motion Maps}
The performance of the multiple stream 3D CNNs, SVM classifiers and average score fusion across the different classifiers are now demonstrated for different lengths of the region adaptive DMM (RA-DMM) templates on the  MSR 3D action dataset. As before, these constitute the RA-DMM for multiple time resolutions to form the RAMDMM.
Table \ref{scratchmodelmsr} includes the results of the recognition model based on RADMM and RAMDMM information templates for different temporal windows.

\begin{table}[th]
\centering
\caption{Performance based on different lengths of RADMM, RAMDMM, average score fusion in terms of MSR 3D Action dataset. \label{scratchmodelmsr}}
\begin{tabular}{|c|c|c|c|c|c|}
 \hline
  {Subsets} & {Scheme} & {5} & {10} & {All} & {RAMDMM}\\
 \hline
 \multirow{3}{*}{AS1}&1/3 &72.10&79.34&95.32&96.53\\&2/3 &78.33&85.49&96.19&98.70\\&Cross &62.95&63.87&85.50&88.31\\
   \hline 
 \multirow{3}{*}{AS2}&1/3 &74.42&77.13&94.11&95.90\\&2/3 &76.16&80.21&95.86&96.91\\&Cross &57.39&62.04&80.27&83.89\\
 \hline
 \multirow{3}{*}{AS3}&1/3 &76.89&81.56&95.87&97.20\\&2/3 &80.29&84.98&96.92&98.38\\&Cross &63.70&66.87&87.16&90.42\\
\hline
\end{tabular}
\end{table}
The results in Table \ref{scratchmodelmsr} show that the recognition performance using RADMM information to form RAMDMM then learning actions' features based on RAMDMM is better than using either traditional DMM or individual length of RADMM. Moreover, it appears to show that sharing a variety of information available from the features by average score fusion between different models can improve the performance of the recognition system. 

\subsubsection{Combining RAMDMM, Multiple Views and Depth Based Multiple Sequences}
Table \ref{msrmultiview} shows the effects of the multi-view RAMDMM (MV-RAMDMM) templates and the effect of the multi-resolution spatio-temporal information on the recognition accuracy of the system also combined with the depth sequences investigated in section \ref{msrdepthsequences}.
\begin{table}[th]
\centering
\caption{Performance of the recognition model based on RADMM, MV-RAMDMM, average fusion of MV-RAMDMM and MR-RGB C3D models in terms of MSR 3D Action dataset. \label{msrmultiview}}
\begin{adjustbox}{max width= 0.49 \textwidth}
\begin{tabular}{|c|c|c|c|c|}
 \hline
  {Subsets} & {Scheme} & {\tiny RAMDMM} & {\tiny MV-RAMDMM} &{\tiny MV-(RAMDMM+depth)}\\
 \hline
 \multirow{3}{*}{AS1}&1/3 &96.53&97.90&99.21\\&2/3 &97.86&98.70&99.91\\&Cross &88.31&91.28&97.95\\

   \hline 

 \multirow{3}{*}{AS2}&1/3 &95.90&97.40&99.08\\&2/3 &96.91&97.89&99.94\\&Cross &83.89&89.11&95.89\\
 
 \hline
 
 \multirow{3}{*}{AS3}&1/3 &97.20&98.33&99.89\\&2/3 &98.21&98.76&99.96\\&Cross &90.42&94.80&95.77\\

\hline
\end{tabular}
\end{adjustbox}
\end{table}

Figure \ref{MSRconfusionAS1AS2AS32nd} shows the confusion matrices of the recognition system using the proposed models under cross-subject evaluation schemes in terms of AS1, AS2 and AS3 subsets of MSR 3D action dataset.

\begin{figure*}[!htb]
\centering
\newcommand\items{8}   
\begin{adjustbox}{max width= 1 \textwidth}
\begin{tabular}{cc*{\items}{|E}|}
\arrayrulecolor{white}
\multicolumn{1}{c}{} &\multicolumn{1}{c}{} &\multicolumn{\items}{c}{} \\ \hhline{~*\items{|-}|}
\multicolumn{1}{c}{} & 
\multicolumn{1}{c}{} & 
\multicolumn{1}{c}{\rot{\textbf{Bend}}} & 
\multicolumn{1}{c}{\rot{\textbf{Fwd.punch}}} & 
\multicolumn{1}{c}{\rot{\textbf{Hammer}}}&
\multicolumn{1}{c}{\rot{\textbf{Hand clap}}} & 
\multicolumn{1}{c}{\rot{\textbf{High throw}}} & 
\multicolumn{1}{c}{\rot{\textbf{Horiz.arm wave}}}&
\multicolumn{1}{c}{\rot{\textbf{Pick-up\& throw}}} & 
\multicolumn{1}{c}{\rot{\textbf{Tennis serve}}}\\
\hhline{~*\items{|-}|}
\multirow{\items}{*}{\rotatebox{90}{}} 
&\textbf{Bend}  & 100 & 0 & 0 & 0 & 0 & 0 & 0 & 0 \\ \hhline{~*\items{|-}|}
&\textbf{Fwd.punch}  & 0 & 100 & 0 & 0 & 0 & 0 & 0 & 0  \\ \hhline{~*\items{|-}|}
&\textbf{Hammer}  & 0 & 16.4 & 83.6 & 0 & 0 & 0 & 0 & 0 \\ \hhline{~*\items{|-}|}
&\textbf{Hand clap}  & 0 & 0  & 0 & 100 & 0 & 0 & 0 & 0   \\ \hhline{~*\items{|-}|}
&\textbf{High throw}  & 0 & 0 & 0 & 0 & 100 & 0 & 0 & 0  \\ \hhline{~*\items{|-}|}
&\textbf{Horiz.arm wave}  & 0 & 0  & 0 & 0 & 0 & 100 & 0 & 0  \\ \hhline{~*\items{|-}|}
&\textbf{Pick-up\&throw}  & 0 & 0  & 0 & 0 & 0 & 0 & 100 & 0  \\ \hhline{~*\items{|-}|}
&\textbf{Tennis serve}  &  0 & 0  & 0 & 0 & 0 & 0 & 0 & 100 \\ \hhline{~*\items{|-}|}
\end{tabular}
\begin{tabular}{cc*{\items}{|E}|}
\arrayrulecolor{white}
\multicolumn{1}{c}{} &\multicolumn{1}{c}{} &\multicolumn{\items}{c}{} \\ \hhline{~*\items{|-}|}
\multicolumn{1}{c}{} & 
\multicolumn{1}{c}{} & 
\multicolumn{1}{c}{\rot{\textbf{Draw circle}}} & 
\multicolumn{1}{c}{\rot{\textbf{Draw cross}}} & 
\multicolumn{1}{c}{\rot{\textbf{Draw tick}}}&
\multicolumn{1}{c}{\rot{\textbf{Forward kick}}} & 
\multicolumn{1}{c}{\rot{\textbf{Hand catch}}} & 
\multicolumn{1}{c}{\rot{\textbf{High arm wave}}}&
\multicolumn{1}{c}{\rot{\textbf{Side-boxing}}} & 
\multicolumn{1}{c}{\rot{\textbf{Two-hand wave}}}\\
\hhline{~*\items{|-}|}
\multirow{\items}{*}{\rotatebox{90}{}} 
&\textbf{Draw circle}  & 100 & 0 & 0 & 0 & 0 & 0 & 0 & 0 \\ \hhline{~*\items{|-}|}
&\textbf{Draw cross}  & 4 & 85 & 0 & 0 & 11 & 0 & 0 & 0  \\ \hhline{~*\items{|-}|}
&\textbf{Draw tick}  & 0 & 0 & 92 & 0 & 5 & 0 & 1 & 0 \\ \hhline{~*\items{|-}|}
&\textbf{Forward kick}  & 0 & 0  & 0 & 100 & 0 & 0 & 0 & 0   \\ \hhline{~*\items{|-}|}
&\textbf{Hand catch}  & 0 & 0 & 8 & 0 & 90 & 0 & 0 & 0  \\ \hhline{~*\items{|-}|}
&\textbf{High arm wave}  & 0 & 0  & 0 & 0 & 0 & 100 & 0 & 0  \\ \hhline{~*\items{|-}|}
&\textbf{Side-boxing}  & 0 & 0  & 0 & 0 & 0 & 0 & 100 & 0  \\ \hhline{~*\items{|-}|}
&\textbf{Two-hand wave}  &  0 & 0  & 0 & 0 & 0 & 0 & 0 & 100 \\ \hhline{~*\items{|-}|}
\end{tabular}
\begin{tabular}{cc*{\items}{|E}|}
\arrayrulecolor{white}
\multicolumn{1}{c}{} &\multicolumn{1}{c}{} &\multicolumn{\items}{c}{} \\ \hhline{~*\items{|-}|}
\multicolumn{1}{c}{} & 
\multicolumn{1}{c}{} & 
\multicolumn{1}{c}{\rot{\textbf{Forward kick}}} & 
\multicolumn{1}{c}{\rot{\textbf{Golf swing}}} & 
\multicolumn{1}{c}{\rot{\textbf{High throw}}}&
\multicolumn{1}{c}{\rot{\textbf{Jogging}}} & 
\multicolumn{1}{c}{\rot{\textbf{Pick-up\&throw}}} & 
\multicolumn{1}{c}{\rot{\textbf{Side kick}}}&
\multicolumn{1}{c}{\rot{\textbf{Tennis serve}}} & 
\multicolumn{1}{c}{\rot{\textbf{Tennis swing}}}\\
\hhline{~*\items{|-}|}
\multirow{\items}{*}{\rotatebox{90}{}} 
&\textbf{Forward kick}  & 100 & 0 & 0 & 0 & 0 & 0 & 0 & 0 \\ \hhline{~*\items{|-}|}
&\textbf{Golf swing}  & 0 & 92 & 0 & 0 & 0 & 1 & 5 & 2  \\ \hhline{~*\items{|-}|}
&\textbf{High throw}  & 0 & 0 & 100 & 0 & 0 & 0 & 0 & 0 \\ \hhline{~*\items{|-}|}
&\textbf{Jogging}  & 0 & 0  & 0 & 100 & 0 & 0 & 0 & 0   \\ \hhline{~*\items{|-}|}
&\textbf{Pick-up\&throw}  & 0 & 0 & 0 & 0 & 97 & 0 & 3 & 0  \\ \hhline{~*\items{|-}|}
&\textbf{Side kick}  & 0 & 0  & 0 & 0 & 0 & 100 & 0 & 0  \\ \hhline{~*\items{|-}|}
&\textbf{Tennis serve}  & 0 & 0  & 0 & 0 & 0 & 0 & 100 & 0  \\ \hhline{~*\items{|-}|}
&\textbf{Tennis swing}  &  0 & 0 & 5 & 0 & 3 & 1 & 14 & 77 \\ \hhline{~*\items{|-}|}
\end{tabular}
\end{adjustbox}
\caption{Confusion matrices of the proposed method, using CS validation scheme in terms of AS1 (left), AS2 (middle) and AS3 (right)  subset of MSR 3D action dataset.}
\label{MSRconfusionAS1AS2AS32nd}
\end{figure*}

Further, a comparison between the proposed method and the state-of-the-art approaches for human action recognition is presented in Table \ref{comparisonMSR2nd} in terms of the MSR Action 3D dataset under the aforementioned evaluation schemes.
\begin{table*}[!htb]
\centering
\caption{Performance of the proposed method compared to the state-of-the-art approaches in terms of the MSR action 3D dataset \cite{li2010action}. 
\label{comparisonMSR2nd}}
\begin{tabular}{|p{2.8cm}|p{0.68cm}|p{0.68cm}|p{0.68cm}|p{0.68cm}|p{0.68cm}|p{0.68cm}|p{0.68cm}|p{0.68cm}|p{0.68cm}|p{0.68cm}|p{0.68cm}|p{0.68cm}|}
 \hline
  \multirow{3}{*}{Method} & \multicolumn{12}{c|}{Accuracy \%}\\ 
  
  &\multicolumn{4}{c|}{1/3 Scheme} & \multicolumn{4}{c|}{2/3 scheme} & \multicolumn{4}{c|}{Cross subject scheme}\\

  &AS1&AS2&AS3&Av.&AS1&AS2&AS3&Av.&AS1&AS2&AS3&Av.\\
  
 \hline
 
 Li et al. \cite{li2010action} &89.5&89.0&96.3&91.6&93.4&92.9&96.3&94.2&71.9&72.9&79.2&74.7\\
 
 DMM-HOG \cite{yang2012recognizing} &97.3&92.2&98.0&95.8&98.7&94.7&98.7&97.4&96.2&84.1&94.6&91.6\\
 
 Chen et al. \cite{chen2016real} &97.3&96.1&98.7&97.4&98.6&98.7&\bf{100}&99.1&96.2&83.2&92.0&90.5\\
 
 HOJ3D \cite{xia2012view} &98.5&96.7&93.5&96.2&98.6&97.2&94.9&97.2&88.0&85.5&63.6&79.0\\
 
 Chaaraoui et al. \cite{chaaraoui2014evolutionary} &-&-&-&-&-&-&-&-&91.6&90.8&97.3&93.2\\
 
 DMM-HOG-KECA \cite{el2015human}  & -& -&-& -& -& -&-& -& 90.6& 90.7&\bf{99.1}&93.5\\
 
 Vemulapalli et al. \cite{vemulapalli2014human} &-&-&-&-&-&-&-&-&95.3&83.9&98.2&92.5\\
 
 STOP \cite{vieira2014improvement} &98.2&94.8&97.4&96.8&99.1&97.0&98.7&98.3&91.7&72.2&98.6&87.5\\
 
 DMM-LBP-FF \cite{chen2015action}&96.7&\bf{100}&99.3&98.7&\bf{100}&\bf{100}&\bf{100}&\bf{100}&98.1&92.0&94.6&94.9\\
 
 DMM-LBP-DF\cite{chen2015action}&98.0&97.4&99.3&98.2&\bf{100}&\bf{100}&\bf{100}&\bf{100}&\bf{99.1}&92.9&92.0&94.7\\
 
 tLDS \cite{ding2018tensor} & - & -& -& -& -& -& -& -&96.81&	89.14&	98.83&	94.85\\
 
 \textbf{Ours} & \textbf{99.2} & \textbf{99.1} & \textbf{99.8} & \textbf{99.3} & \textbf{99.9} & \textbf{99.9} & \textbf{99.8} & \textbf{99.9} & \textbf{97.9} & \textbf{95.8} & \textbf{95.7} & \textbf{96.5}\\ 
 
\hline
\end{tabular}
\end{table*}

It can be seen that our method outperforms the state-of-the-art approaches for the majority of cases and in others achieves at least comparable performance. Even though some of them are DMM based methods such as \cite{chen2015action} and \cite{yang2012recognizing}, our method achieves greater recognition rate in the range of 1-6\%. This appears to indicate that MV-RAMDMM and spatio-temporal information based features can provide more powerful discrimination. Our approach utilises adaptive multiple hierarchical features that cover various periods of an action. In addition, the pre-trained recognition model uses a diverse range of layers which improves the chances to obtain the most accurate recognition performance.  

\subsection{MSR 3D Daily Activity}

The Microsoft Research (MSR) daily activity 3D dataset is among the most challenging datasets because of a high level of intra-class variation and many of the actions are based on object interaction. An action with object interaction is where the subject is interacting with an object when performing an action. This dataset has been captured by a Kinect sensor. It consists of depth and RGB sequences and includes 16 activities: drink, eat, read book, call cellphone,write on a paper, use laptop, use vacuum cleaner, cheer up, sit still, toss paper, play game, lay down on sofa, walk, play guitar, stand up, sit down. Performed by 10 subjects each subject performs an action twice in two different poses (standing and sitting). 

Different evaluation schemes have been considered in the literature in terms of MSR daily activity 3D dataset. Here, similar to \cite{wang2016action}, a cross-subject validation is performed with subjects 1,3,5,7,9 for training and subjects 2,4,6,8,10 for testing.
Here the person and pose detection steps are used to detect and localise a person within a frame and pose detection is used to identify the pose, whether sitting or standing.

\subsubsection{Multi-Resolution in Time Appearance Information}
Firstly, multiple temporal resolutions $(10, 16, 25)$ of RGB information are investigated separately with the fine-tuned C3D models. The outputs of these models are, as usual, classified using different SVMs. As before, the SVM outputs are combined using average score fusion. The results for this purely multi-temporal appearance based recognition sub-system are shown in Table \ref{C3Ddaily2nd}. 
\begin{table}[ht]
\centering
\caption{Results of using multi-temporal resolution RGB data for the MSR 3D daily activity dataset. \label{C3Ddaily2nd}}
\begin{tabular}{|c|c|c|c|c|c|c|c|c|}
    \hline
  &   {RGB$_{10}$} & {RGB$_{16}$} & {RGB$_{25}$} & {RGB$_{All}$}\\ 
    \hline
Sit   & 53.73&57.50&64.48&65.90\\
Stand & 51.20&56.81&61.11&63.79\\
    \hline
\end{tabular}
\end{table}
It can be seen in Table \ref{C3Ddaily2nd}, that, as before, the robustness of the proposed model improves with an increase in the number of frames included in the system with the best combining the results from all temporal resolutions. This dataset is often considered to be much more complicated than others due to the two different scenarios for each single action; but the hierarchical strategy with the fine-tuned model is able to achieve comparable results based on RGB raw data. Moreover, a reasonable overall performance is also achieved that reaches 64.85\% when an average recognition rate is employed.

\subsubsection{Multi-Resolution in Time Region Adaptive Depth Motion Maps}
As before, the RADMMs templates for three different temporal windows are computed and fed into fine-tuned C3D models, multi-class SVMs the results of which constitute the RAMDMM for action recognition. Competitive results are achieved using these improved multiple temporal resolutions as can be seen in Table \ref{C3DdailyDMM12nd}

\begin{table} [th]
\centering
\caption{Results of RADMM, RAMDMM, and MV-RMDMM with MR-RGB in terms of sitting pose of the MSR 3D daily activity dataset. \label{C3DdailyDMM12nd}}
\begin{adjustbox}{max width= 0.49 \textwidth}
\begin{tabular}{|c|c|c|c|c|c|}
 \hline
 {5} & {10} & {All} & {\tiny RAMDMM} &{\tiny MV-RAMDMM}& {\tiny MV-(RMDMM + RGB)} \\
 \hline
 65.19&70.57&79.90&81.32&87.76&89.00\\

\hline 
\end{tabular}
\end{adjustbox}
\end{table}

\begin{table} [th]
\centering
\caption{Results of RADMM, RAMDMM, and MV-RMDMM with MR-RGB in terms of standing pose of the MSR 3D daily activity dataset. \label{C3DdailyDMM22nd}}
\begin{adjustbox}{max width= 0.49 \textwidth}
\begin{tabular}{|c|c|c|c|c|c|}
 \hline
 {5} & {10} & {All} & {\tiny RAMDMM} &{\tiny MV-RAMDMM}& {\tiny MV-RMDMM+RGB} \\
 \hline
 64.92&68.70&77.18&78.66&83.53&86.00\\
\hline 
\end{tabular}
\end{adjustbox}
\end{table}

For multiple views (MV-RAMDMM) the performance reaches (89.00\%) and (86.00\%) within sitting and standing poses as presented in Table \ref{C3DdailyDMM22nd}.

\begin{figure*}[!htb]
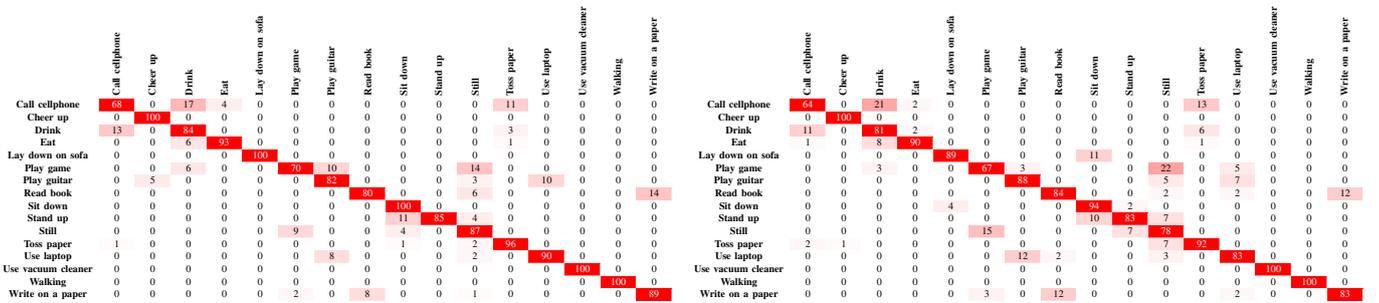

\centering
\newcommand\items{16}   
\begin{adjustbox}{max width= 1 \textwidth}
\begin{tabular}{cc*{\items}{|E}|}
\arrayrulecolor{white}
\multicolumn{1}{c}{} &\multicolumn{1}{c}{} &\multicolumn{\items}{c}{} \\ \hhline{~*\items{|-}|}
\multicolumn{1}{c}{} & 
\multicolumn{1}{c}{} & 
\multicolumn{1}{c}{\rot{\textbf{Call cellphone}}} & 
\multicolumn{1}{c}{\rot{\textbf{Cheer up}}} & 
\multicolumn{1}{c}{\rot{\textbf{Drink}}} & 
\multicolumn{1}{c}{\rot{\textbf{Eat}}}&
\multicolumn{1}{c}{\rot{\textbf{Lay down on sofa}}} & 
\multicolumn{1}{c}{\rot{\textbf{Play game}}}&
\multicolumn{1}{c}{\rot{\textbf{Play guitar}}} & 
\multicolumn{1}{c}{\rot{\textbf{Read book}}}&
\multicolumn{1}{c}{\rot{\textbf{Sit down}}}&
\multicolumn{1}{c}{\rot{\textbf{Stand up}}} & 
\multicolumn{1}{c}{\rot{\textbf{Still}}} & 
\multicolumn{1}{c}{\rot{\textbf{Toss paper}}}&
\multicolumn{1}{c}{\rot{\textbf{Use laptop}}} & 
\multicolumn{1}{c}{\rot{\textbf{Use vacuum cleaner}}}&
\multicolumn{1}{c}{\rot{\textbf{Walking}}}&
\multicolumn{1}{c}{\rot{\textbf{Write on a paper}}}\\
\hhline{~*\items{|-}|}
\multirow{\items}{*}{\rotatebox{90}{}} 
&\textbf{Call cellphone} & 68 & 0 & 17 & 4 & 0 & 0 & 0 & 0&0&0& 0 & 11 & 0 & 0&0&0 \\ \hhline{~*\items{|-}|}
&\textbf{Cheer up}  & 0 & 100 & 0 & 0 & 0 & 0 & 0 & 0 &0&0& 0 & 0 & 0 & 0&0&0 \\ \hhline{~*\items{|-}|}
&\textbf{Drink}  & 13 & 0 & 84 & 0 & 0 & 0 & 0 & 0 &0&0& 0 & 3 & 0 & 0&0&0\\ \hhline{~*\items{|-}|}
&\textbf{Eat}  & 0 & 0  & 6& 93 & 0 & 0 & 0 & 0&0&0 & 0 & 1 & 0 & 0&0&0  \\ \hhline{~*\items{|-}|}
&\textbf{Lay down on sofa}  & 0 & 0 & 0 & 0 & 100 & 0 & 0 & 0&0&0 & 0 & 0 & 0 & 0&0&0 \\ \hhline{~*\items{|-}|}
&\textbf{Play game}  & 0 & 0 & 6 & 0 & 0 & 70 & 10 & 0&0&0 & 14 & 0 & 0 & 0&0&0 \\ \hhline{~*\items{|-}|}
&\textbf{Play guitar}  & 0 & 5  & 0 & 0 & 0 & 0 & 82 & 0&0&0 & 3 & 0 & 10 & 0&0&0 \\ \hhline{~*\items{|-}|}
&\textbf{Read book}  &  0 & 0  & 0 & 0 & 0 & 0 & 0 & 80&0&0& 6 & 0 & 0 & 0&0&14 \\ \hhline{~*\items{|-}|}
&\textbf{Sit down}  &  0 & 0 & 0 & 0 & 0 & 0 & 0 & 0&100& 0 & 0 & 0 & 0 & 0&0&0\\ \hhline{~*\items{|-}|}
&\textbf{Stand up}  & 0 & 0 & 0 & 0 & 0 & 0 & 0 & 0&11&85& 4 & 0 & 0 & 0&0&0 \\ \hhline{~*\items{|-}|}
&\textbf{Still}  & 0 & 0 & 0 & 0 & 0 & 9 & 0 & 0&4&0& 87 & 0 & 0 & 0&0&0 \\ \hhline{~*\items{|-}|}
&\textbf{Toss paper}  & 1 & 0 & 0 & 0 & 0 & 0 & 0 & 0&1&0 & 2 & 96 & 0 & 0&0&0 \\ \hhline{~*\items{|-}|}
&\textbf{Use laptop}  & 0 & 0 & 0 & 0 & 0 & 0 & 8 & 0&0&0 & 2 & 0 & 90 & 0&0&0 \\ \hhline{~*\items{|-}|}
&\textbf{Use vacuum cleaner}  & 0 & 0  & 0 & 0 & 0 & 0 & 0 & 0&0&0 & 0 & 0 & 0 & 100&0&0 \\ \hhline{~*\items{|-}|}
&\textbf{Walking}  &  0 & 0  & 0 & 0 & 0 & 0 & 0 & 0&0&0 & 0 & 0 & 0 & 0&100&0\\ \hhline{~*\items{|-}|}
&\textbf{Write on a paper}  &  0 & 0 & 0 & 0 & 0 & 2 & 0 & 8&0&0& 1 & 0 & 0 & 0&0&89 \\ \hhline{~*\items{|-}|}
\end{tabular}

\begin{tabular}{cc*{\items}{|E}|}
\arrayrulecolor{white}
\multicolumn{1}{c}{} &\multicolumn{1}{c}{} &\multicolumn{\items}{c}{} \\ \hhline{~*\items{|-}|}
\multicolumn{1}{c}{} & 
\multicolumn{1}{c}{} & 
\multicolumn{1}{c}{\rot{\textbf{Call cellphone}}} & 
\multicolumn{1}{c}{\rot{\textbf{Cheer up}}} & 
\multicolumn{1}{c}{\rot{\textbf{Drink}}} & 
\multicolumn{1}{c}{\rot{\textbf{Eat}}}&
\multicolumn{1}{c}{\rot{\textbf{Lay down on sofa}}} & 
\multicolumn{1}{c}{\rot{\textbf{Play game}}}&
\multicolumn{1}{c}{\rot{\textbf{Play guitar}}} & 
\multicolumn{1}{c}{\rot{\textbf{Read book}}}&
\multicolumn{1}{c}{\rot{\textbf{Sit down}}}&
\multicolumn{1}{c}{\rot{\textbf{Stand up}}} & 
\multicolumn{1}{c}{\rot{\textbf{Still}}} & 
\multicolumn{1}{c}{\rot{\textbf{Toss paper}}}&
\multicolumn{1}{c}{\rot{\textbf{Use laptop}}} & 
\multicolumn{1}{c}{\rot{\textbf{Use vacuum cleaner}}}&
\multicolumn{1}{c}{\rot{\textbf{Walking}}}&
\multicolumn{1}{c}{\rot{\textbf{Write on a paper}}}\\
\hhline{~*\items{|-}|}
\multirow{\items}{*}{\rotatebox{90}{}} 
&\textbf{Call cellphone} & 64 & 0 & 21 & 2 & 0 & 0 & 0 & 0&0&0& 0 & 13 & 0 & 0&0&0 \\ \hhline{~*\items{|-}|}
&\textbf{Cheer up}  & 0 & 100 & 0 & 0 & 0 & 0 & 0 & 0 &0&0& 0 & 0 & 0 & 0&0&0 \\ \hhline{~*\items{|-}|}
&\textbf{Drink}  & 11 & 0 & 81 & 2 & 0 & 0 & 0 & 0 &0&0& 0 & 6 & 0 & 0&0&0\\ \hhline{~*\items{|-}|}
&\textbf{Eat}  & 1 & 0  & 8& 90 & 0 & 0 & 0 & 0&0&0 & 0 & 1 & 0 & 0&0&0  \\ \hhline{~*\items{|-}|}
&\textbf{Lay down on sofa}  & 0 & 0 & 0 & 0 & 89 & 0 & 0 & 0&11&0 & 0 & 0 & 0 & 0&0&0 \\ \hhline{~*\items{|-}|}
&\textbf{Play game}  & 0 & 0 & 3 & 0 & 0 & 67 & 3 & 0&0&0 & 22 & 0 & 5 & 0&0&0 \\ \hhline{~*\items{|-}|}
&\textbf{Play guitar}  & 0 & 0  & 0 & 0 & 0 & 0 & 88 & 0&0&0 & 5 & 0 & 7 & 0&0&0 \\ \hhline{~*\items{|-}|}
&\textbf{Read book}  &  0 & 0  & 0 & 0 & 0 & 0 & 0 & 84&0&0& 2 & 0 & 2 & 0&0&12 \\ \hhline{~*\items{|-}|}
&\textbf{Sit down}  &  0 & 0 & 0 & 0 & 4 & 0 & 0 & 0&94&2 & 0 & 0 & 0 & 0&0&0\\ \hhline{~*\items{|-}|}
&\textbf{Stand up}  & 0 & 0 & 0 & 0 & 0 & 0 & 0 & 0&10&83& 7 & 0 & 0 & 0&0&0 \\ \hhline{~*\items{|-}|}
&\textbf{Still}  & 0 & 0 & 0 & 0 & 0 & 15 & 0 & 0&0&7& 78 & 0 & 0 & 0&0&0 \\ \hhline{~*\items{|-}|}
&\textbf{Toss paper}  & 2 & 1 & 0 & 0 & 0 & 0 & 0 & 0&0&0 & 7 & 92 & 0 & 0&0&0 \\ \hhline{~*\items{|-}|}
&\textbf{Use laptop}  & 0 & 0 & 0 & 0 & 0 & 0 & 12 & 2&0&0 & 3 & 0 & 83 & 0&0&0 \\ \hhline{~*\items{|-}|}
&\textbf{Use vacuum cleaner}  & 0 & 0  & 0 & 0 & 0 & 0 & 0 & 0&0&0 & 0 & 0 & 0 & 100&0&0 \\ \hhline{~*\items{|-}|}
&\textbf{Walking}  &  0 & 0  & 0 & 0 & 0 & 0 & 0 & 0&0&0 & 0 & 0 & 0 & 0&100&0\\ \hhline{~*\items{|-}|}
&\textbf{Write on a paper}  &  0 & 0 & 0 & 0 & 0 & 3 & 0 & 12&0&0& 0 & 0 & 2 & 0&0&83 \\ \hhline{~*\items{|-}|}
\end{tabular}
\end{adjustbox}
\caption{Confusion matrices of the proposed method through standing pose (right) and sitting pose (left) in terms of daily activity 3D dataset.}
\label{Dailyconfusion2nd2}
\end{figure*}

Further improvements can be seen by involving the multi-resolution spatio-temporal RGB information. Average score fusion improves the recognition of some objects-interaction actions and accomplishes (89\%) and (86\%) in terms of sitting and standing poses respectively. The overall recognition rate of all datasets can be calculated taking the average of the two poses recognition rates which reaches (87.5\%). Figs. \ref{Dailyconfusion2nd2} shows the confusion matrix of the hierarchical recognition system in terms of MSR 3D daily activity dataset.

A comparison between the proposed method and state-of-the-art approaches for action recognition is introduced in Table \ref{dailycomparison2} in terms of MSR 3D daily activity dataset.

\begin{table}[th]
\centering
\caption{Comparison of our method with the state-of-the-art approaches in terms of MSR daily activity 3D dataset \cite{wang2012mining}. \label{dailycomparison2}}
\begin{tabular}{|c|c|}
 \hline
 {Method} & {Accuracy \%}\\
 \hline
 LOP \cite{wang2012mining} & 42.50\\
 Depth Motion Maps \cite{yang2012recognizing}&43.13\\
 Local HON4D \cite{oreifej2013hon4d}&80.00\\
 Actionlet Ensemble \cite{wang2012mining} & 85.75\\
 SNV \cite{yang2017super}& 86.25\\
 Range Sample \cite{lu2014range}& 95.63\\
 DMM-CNN \cite{wang2016action}& 85.00\\
 \textbf{Ours} & \textbf{87.50}\\ 
\hline
\end{tabular}
\end{table}

In Table \ref{dailycomparison2}, it can be seen that limited accuracy was previously achieved by LOP \cite{wang2012mining} and DMM \cite{yang2012recognizing} based approaches. Local HON4D was designed in \cite{oreifej2013hon4d} to tackle this kind of limitation and achieved a recognition rate of 80.00\%. Actionlet Ensemble in \cite{wang2012mining} and SNV in \cite{yang2017super} achieved a recognition rate that reaches 85.75\% and 86.25\% respectively. These used a combination of depth and skeleton data. A recent method in \cite{wang2016action} indicated the importance of DMM information and suggested the use of Temporal Depth Motion Maps and fine-tuned convolutional models. It achieved a relatively competitive result of 85.00\%. Our method achieves comparable results with an improvement over some methods using our MV-RAMDMM and the spatio-temporal information of the C3D model. However, our method performed worse than the Range Sample \cite{lu2014range} technique. This can be explained due to the noisy, complex and dynamic background
of this dataset which can introduce significant noise in the RAMDMMs. Moreover, the Range Sample \cite{lu2014range} method contained a technique that used skeleton data to eliminate the noise from the background. The confusion matrices in terms of MSR daily activity 3D dataset are shown in Fig. \ref{Dailyconfusion2nd2}.

\section{Conclusions}

A novel feature representation technique for RGB-D data has been presented that enables multi-view and multi-temporal action recognition. A Multiple view and Multi-resolution Region Adaptive Depth Motion Maps (RA-DMMs) representation is proposed. The different views include the original and synthesised view-points to achieve view-invariant recognition. This work also makes use of temporal motion information more effectively. It integrates it into the depth sequences to help build in, by design, invariance to variations in an action's speed. An adaptive weighting approach is employed to help differentiate between the most important stages of an action. Appearance information in terms of multi-temporal RGB data is used to help retain a focus on the underlying appearance information that would otherwise be lost with depth data alone. This helps to provide sensitivity to interactions with small objects. Compact and discriminative spatio-temporal features are extracted using a series of fine-tuned 3D Convolutional Neural Networks (3D CNN)s. In addition, a pose estimation system is employed to achieve a hierarchical recognition structure. This helps the model to recognise the same action but with different positions. Multi-class Support Vector Machines (SVM)s are used for action classification. Then, late score fusion technique is employed between different streams for the final decision.

The proposed method is robust enough to recognise human activities even with small differences in actions. This is in addition to achieving improved performance that is invariant to multiple view-points and providing excellent performance on actions that partly depend on human-object interactions. The system also  remains invariant to a noisy environment and errors in the depth maps and temporal misalignments.

The proposed approach has been extensively validated on three benchmark datasets: MSR 3D actions, Northwestern UCLA multi-view actions and MSR daily activities. The experimental results have demonstrated the great performance of the proposed method in comparison to state-of-the-art approaches.

\bibliographystyle{IEEEtran}
\bibliography{sample}

\end{document}